\documentclass[letterpaper, 10 pt, conference]{ieeeconf}
\IEEEoverridecommandlockouts
\usepackage{times}

\makeatletter
\let\NAT@parse\undefined
\makeatother
\usepackage[numbers]{natbib}
\usepackage{multicol}
\usepackage[symbols,nogroupskip,nonumberlist,stylemods={super},style=super]{glossaries-extra}
\makeglossaries
\usepackage[bookmarks=true]{hyperref}
\usepackage{siunitx}
\usepackage{tabularx}
\usepackage{graphicx}
\graphicspath{ {./figures/} }
\usepackage{subcaption}
\usepackage{amsmath}
\usepackage{amssymb}
\usepackage{acro}
\usepackage{listings}
\usepackage{xcolor}
\usepackage{makecell}
\usepackage{multirow}
\usepackage{colortbl}
\definecolor{codegreen}{rgb}{0,0.6,0}
\definecolor{codegray}{rgb}{0.5,0.5,0.5}
\definecolor{lightgray}{rgb}{0.75,0.75,0.75}
\definecolor{codepurple}{rgb}{0.58,0,0.82}
\definecolor{backcolour}{rgb}{0.95,0.95,0.92}
\lstdefinestyle{mystyle}{
    backgroundcolor=\color{backcolour},   
    commentstyle=\color{codegreen},
    keywordstyle=\color{magenta},
    numberstyle=\tiny\color{codegray},
    stringstyle=\color{codepurple},
    basicstyle=\ttfamily\scriptsize,
    breakatwhitespace=false,         
    breaklines=true,                 
    captionpos=b,                    
    keepspaces=true,                 
    numbers=left,                    
    numbersep=5pt,                  
    showspaces=false,                
    showstringspaces=false,
    showtabs=false,                  
    tabsize=2
}
\let\labelindent\relax 
\usepackage{enumitem}
\usepackage{comment}
\usepackage[multiple]{footmisc}
\let\oldFootnote\footnote
\newcommand\nextToken\relax
\usepackage{eso-pic}
 \AddToShipoutPicture*{\AtPageUpperLeft{\small \sffamily\raisebox{-1.2cm}{\hspace{1.8cm}\parbox{\textwidth}{\copyright 2024 IEEE.  Personal use of this material is permitted.  Permission from IEEE must be obtained for all other uses, in any current or future media, including reprinting/republishing this material for advertising or promotional purposes, creating new collective works, for resale or redistribution to servers or lists, or reuse of any copyrighted component of this work in other works.}}}}

\renewcommand\footnote[1]{%
    \oldFootnote{#1}\futurelet\nextToken\isFootnote}

\newcommand\isFootnote{%
    \ifx\footnote\nextToken\textsuperscript{,}\fi}
\DeclareAcronym{UAV}{
  short = UAV,
  long  = uncrewed aerial vehicle,
  short-indefinite = a,
  long-indefinite = an
}

\DeclareAcronym{IMU}{
  short = IMU,
  long = inertial measurement unit,
  short-indefinite = an,
  long-indefinite = an
}

\DeclareAcronym{FMCW}{
  short = FMCW,
  long = frequency-modulated continuous-wave,
  short-indefinite = an,
  long-indefinite = a
}

\DeclareAcronym{RIO}{
  short = RIO,
  long = radar-inertial odometry,
  short-indefinite = a,
  long-indefinite = a
}

\DeclareAcronym{BRIO}{
  short = BRIO,
  long = baro-radar-inertial odometry,
  short-indefinite = a,
  long-indefinite = a
}

\DeclareAcronym{CFAR}{
  short = CFAR,
  long = constant false alarm rate,
  short-indefinite = a,
  long-indefinite = a
}

\DeclareAcronym{DoA}{
  short = DoA,
  long = direction-of-arrival,
  short-indefinite = a,
  long-indefinite = a,
  long-plural-form = directions-of-arrival
}

\DeclareAcronym{RANSAC}{
  short = RANSAC,
  long = random sample consensus,
  short-indefinite = a,
  long-indefinite = a
}

\DeclareAcronym{MLESAC}{
  short = MLESAC,
  long = maximum likelihood consensus,
  short-indefinite = an,
  long-indefinite = a
}

\DeclareAcronym{MEMS}{
  short = MEMS,
  long = micro-electromechanical systems,
  short-indefinite = a,
  long-indefinite = a
}

\DeclareAcronym{SLAM}{
  short = SLAM,
  long = simultaneous localization and mapping,
  short-indefinite = a,
  long-indefinite = a
}

\DeclareAcronym{MIMO}{
  short = MIMO,
  long = multiple-input and multiple-output,
  short-indefinite = a,
  long-indefinite = a
}

\DeclareAcronym{ADC}{
  short = ADC,
  long = analog-to-digital converter,
  short-indefinite = an,
  long-indefinite = an
}

\DeclareAcronym{SNR}{
  short = SNR,
  long = signal-to-noise ratio,
  short-indefinite = an,
  long-indefinite = a
}

\DeclareAcronym{MAP}{
  short = MAP,
  long = maximum a posteriori probability,
  short-indefinite = an,
  long-indefinite = a
}

\DeclareAcronym{GNSS}{
  short = GNSS,
  long = global navigation satellite system,
  short-indefinite = a,
  long-indefinite = a
}

\DeclareAcronym{CAD}{
  short = CAD,
  long = computer-aided design,
  short-indefinite = a,
  long-indefinite = a
}

\DeclareAcronym{FOV}{
  short = FOV,
  long = field of view,
  short-indefinite = a,
  long-indefinite = a
}

\DeclareAcronym{RCS}{
  short = RCS,
  long = radar cross section,
  short-indefinite = an,
  long-indefinite = a
}

\DeclareAcronym{VIO}{
  short = VIO,
  long = visual-inertial odometry,
  short-indefinite = a,
  long-indefinite = a
}

\DeclareAcronym{LIO}{
  short = LIO,
  long = lidar-inertial odometry,
  short-indefinite = a,
  long-indefinite = a
}

\DeclareAcronym{RPE}{
  short = RPE,
  long = relative pose error,
  short-indefinite = a,
  long-indefinite = a
}

\DeclareAcronym{RMSE}{
  short = RMSE,
  long = root-mean-square error,
  short-indefinite = a,
  long-indefinite = a
}
\newcommand{\reffig}[1]{Fig.~\ref{#1}}
\newcommand{\reftab}[1]{Table~\ref{#1}}

\newcommand{\refequ}[1]{\eqref{#1}}

\makeatletter
\newcommand{\pushright}[1]{\ifmeasuring@#1\else\omit\hfill$\displaystyle#1$\fi\ignorespaces}
\newcommand{\pushleft}[1]{\ifmeasuring@#1\else\omit$\displaystyle#1$\hfill\fi\ignorespaces}
\makeatother

\newcommand{\rev}[1]{\textcolor{black}{#1}}

\DeclareSIUnit\g{g}
\glsxtrnewsymbol[description={Inertial frame}]{frame_inertial}{\ensuremath{I}}
\glsxtrnewsymbol[description={Inertial frame x-axis}]{frame_inertial_x}{\ensuremath{_I\mathbf{e}_x}}
\glsxtrnewsymbol[description={Inertial frame y-axis}]{frame_inertial_y}{\ensuremath{_I\mathbf{e}_y}}
\glsxtrnewsymbol[description={Inertial frame z-axis}]{frame_inertial_z}{\ensuremath{_I\mathbf{e}_z}}
\glsxtrnewsymbol[description={Body frame}]{frame_body}{\ensuremath{B}}
\glsxtrnewsymbol[description={Body frame x-axis}]{frame_body_x}{\ensuremath{_B\mathbf{e}_x}}
\glsxtrnewsymbol[description={Body frame y-axis}]{frame_body_y}{\ensuremath{_B\mathbf{e}_y}}
\glsxtrnewsymbol[description={Body frame z-axis}]{frame_body_z}{\ensuremath{_B\mathbf{e}_z}}
\glsxtrnewsymbol[description={Radar frame}]{frame_radar}{\ensuremath{R}}
\glsxtrnewsymbol[description={Radar frame x-axis}]{frame_radar_x}{\ensuremath{_R\mathbf{e}_x}}
\glsxtrnewsymbol[description={Radar frame y-axis}]{frame_radar_y}{\ensuremath{_R\mathbf{e}_y}}
\glsxtrnewsymbol[description={Radar frame z-axis\newline}]{frame_radar_z}{\ensuremath{_R\mathbf{e}_z}}

\glsxtrnewsymbol[description={Set of all estimated states up to time $k$}]{set_state}{\ensuremath{\mathcal{X}_k}}
\glsxtrnewsymbol[description={Set of \ac{MAP} state estimates up to time $k$}]{set_state_map}{\ensuremath{\mathcal{X}_k^{\mathrm{MAP}}}}
\glsxtrnewsymbol[description={State vector}]{state_vector}{\ensuremath{\mathbf{x}}}
\glsxtrnewsymbol[description={Robot pose}]{robot_pose}{\ensuremath{\mathbf{T}_{IB}}}
\glsxtrnewsymbol[description={Robot orientation}]{robot_orientation}{\ensuremath{\mathbf{R}_{IB}}}
\glsxtrnewsymbol[description={Robot translation}]{robot_translation}{\ensuremath{_I \mathbf{t}_{IB}}}
\glsxtrnewsymbol[description={Robot linear velocity}]{robot_linear_velocity}{\ensuremath{_I \mathbf{v}_{IB}}}
\glsxtrnewsymbol[description={Radar linear velocity in radar frame}]{vel_radar}{\ensuremath{_R\mathbf{v}_{IR}}}
\glsxtrnewsymbol[description={Radar linear velocity in inertial frame}]{vel_radar_inertial}{\ensuremath{_I\mathbf{v}_{IR}}}
\glsxtrnewsymbol[description={Gyroscope bias}]{bias_gyro}{\ensuremath{_B \mathbf{b}_g}}
\glsxtrnewsymbol[description={Accelerometer bias\newline}]{bias_acc}{\ensuremath{_B \mathbf{b}_a}}

\glsxtrnewsymbol[description={Set of all estimated zero-velocity track positions up to time $k$}]{set_track_estimates}{\ensuremath{\mathcal{T}_k}}
\glsxtrnewsymbol[description={Zero-velocity track}]{track}{\ensuremath{\mathbf{t}}}
\glsxtrnewsymbol[description={Tracked zero-velocity detection position\newline}]{track_position}{\ensuremath{_I\mathbf{t}_{IT}}}

\glsxtrnewsymbol[description={Set of all measurements up to time $k$\newline}]{set_meas}{\ensuremath{\mathcal{Z}_k}}
\glsxtrnewsymbol[description={Set of all radar detections in a frame}]{set_detections}{\ensuremath{\mathcal{R}}}
\glsxtrnewsymbol[description={Radar detection}]{radar_detection}{\ensuremath{T}}
\glsxtrnewsymbol[description={Radar detection bearing vector}]{radar_detection_bearing}{\ensuremath{_R\mathbf{e}_{T}}}
\glsxtrnewsymbol[description={Radar detection position vector}]{radar_detection_position}{\ensuremath{{_R}\mathbf{t}_{RT}}}
\glsxtrnewsymbol[description={Radar detection Doppler velocity}]{radar_detection_doppler}{\ensuremath{v_{T}}}
\glsxtrnewsymbol[description={Radar detection \ac{SNR}}]{radar_detection_snr}{\ensuremath{\mathrm{SNR}_{T}}}
\glsxtrnewsymbol[description={Radar detection noise}]{radar_detection_noise}{\ensuremath{w_{T}}}
\glsxtrnewsymbol[description={Radar detection bearing vector bias\newline}]{radar_detection_bearing_bias}{\ensuremath{\alpha}}

\glsxtrnewsymbol[description={Set of all \ac{IMU} measurements between two radar frames}]{set_imu}{\ensuremath{\mathcal{I}}}
\glsxtrnewsymbol[description={Robot angular velocity}]{robot_angular_velocity}{\ensuremath{{_B\boldsymbol{\omega}}_{IB}}}
\glsxtrnewsymbol[description={Robot linear acceleration\newline}]{robot_linear_acceleration}{\ensuremath{{_B}\mathbf{a}_B}}

\glsxtrnewsymbol[description={Set of barometer measurements}]{set_baro}{\ensuremath{\mathcal{B}}}
\glsxtrnewsymbol[description={Barometer height measurement}]{baro_meas}{\ensuremath{z_p}}
\glsxtrnewsymbol[description={Barometer pressure measurement\newline}]{baro_pressure_meas}{\ensuremath{p}}

\glsxtrnewsymbol[description={Set of all radar frames up to time $k$}]{set_radar}{\ensuremath{\mathcal{K}_k}}
\glsxtrnewsymbol[description={Set of all radar frames containing track $j$}]{set_track}{\ensuremath{\mathcal{K}_j}}
\glsxtrnewsymbol[description={Set of all zero-velocity tracks up to time $k$}]{set_tracks}{\ensuremath{\mathcal{L}_k}}
\glsxtrnewsymbol[description={Time index}]{index_time}{\ensuremath{i}}
\glsxtrnewsymbol[description={Zero-velocity track index}]{index_track}{\ensuremath{j}}
\glsxtrnewsymbol[description={Most recent time index}]{index_current_time}{\ensuremath{k}}
\glsxtrnewsymbol[description={Detection in radar frame index\newline}]{index_detection}{\ensuremath{m}}

\glsxtrnewsymbol[description={Robust loss function}]{robust_loss_function}{\ensuremath{\rho}}
\glsxtrnewsymbol[description={\ac{IMU} factor residual}]{residual_imu}{\ensuremath{\mathbf{r}_{I}}}
\glsxtrnewsymbol[description={Zero-velocity track factor residual}]{residual_zero_velocity}{\ensuremath{\mathbf{r}_{T}}}
\glsxtrnewsymbol[description={Doppler factor residual}]{residual_doppler}{\ensuremath{{r}_{D}}}
\glsxtrnewsymbol[description={Barometer factor residual}]{residual_baro}{\ensuremath{{r}_{B}}}
\glsxtrnewsymbol[description={Prior factor residual}]{residual_prior}{\ensuremath{\mathbf{r}_{P}}}
\glsxtrnewsymbol[description={Prior orientation factor residual}]{residual_prior_orientation}{\ensuremath{\mathbf{r}_{\gls{robot_orientation}}}}
\glsxtrnewsymbol[description={Prior position factor residual}]{residual_prior_position}{\ensuremath{\mathbf{r}_{\gls{robot_translation}}}}
\glsxtrnewsymbol[description={Prior linear velocity factor residual}]{residual_prior_velocity}{\ensuremath{\mathbf{r}_{\gls{robot_linear_velocity}}}}
\glsxtrnewsymbol[description={Prior gyro bias factor residual}]{residual_prior_bias_gyro}{\ensuremath{\mathbf{r}_{\gls{bias_gyro}}}}
\glsxtrnewsymbol[description={Prior accelerometer bias factor residual\newline}]{residual_prior_bias_acc}{\ensuremath{\mathbf{r}_{\gls{bias_acc}}}}

\glsxtrnewsymbol[description={Radar extrinsic calibration}]{radar_calibration}{\ensuremath{\mathbf{T}_{BR}}}
\glsxtrnewsymbol[description={Radar calibration rotation}]{radar_calibration_rotation}{\ensuremath{\mathbf{R}_{BR}}}
\glsxtrnewsymbol[description={Radar calibration translation\newline}]{radar_calibration_translation}{\ensuremath{_B\mathbf{t}_{BR}}}

\glsxtrnewsymbol[description={Prior measurement covariance matrix}]{cov_prior}{\ensuremath{\mathbf{\Sigma}_P}}
\glsxtrnewsymbol[description={IMU measurement covariance matrix}]{cov_imu}{\ensuremath{\mathbf{\Sigma}_I}}
\glsxtrnewsymbol[description={Zero-velocity track position measurement covariance matrix}]{cov_track}{\ensuremath{\mathbf{\Sigma}_T}}
\glsxtrnewsymbol[description={Doppler measurement covariance}]{cov_doppler}{\ensuremath{\Sigma_D}}
\glsxtrnewsymbol[description={Barometer measurement covariance}]{cov_baro}{\ensuremath{\Sigma_B}}
\glsxtrnewsymbol[description={Doppler measurement standard deviation}]{std_doppler}{\ensuremath{\sigma_D}}
\glsxtrnewsymbol[description={Barometer measurement standard deviation\newline}]{std_baro}{\ensuremath{\sigma_B}}

\glsxtrnewsymbol[description={Velocity control input}]{control_input}{\ensuremath{\mathbf{v}_\mathrm{ref}}}

\title{\LARGE \bf
A robust baro-radar-inertial odometry m-estimator \\ for multicopter navigation in cities and forests
}

\author{Rik Girod$^{1}$, Marco Hauswirth$^{1}$, Patrick Pfreundschuh$^{1}$, Mariano Biasio$^{1}$, and Roland Siegwart$^{1}$
\thanks{*This work was supported by Armasuisse S+T.}
\thanks{$^{1}$Authors are with Autonomous Systems Lab, ETH Zürich, 8092 Zürich, Switzerland
        {\tt\small \{brik, haumarco, patripfr, mbiasio, rsiegwart\}@ethz.ch}}%
}

\begin{document}
\maketitle
\thispagestyle{empty}
\pagestyle{empty}

\begin{abstract}
Search and rescue operations require mobile robots to navigate unstructured indoor and outdoor environments. 
In particular, actively stabilized multirotor drones need precise movement data to balance and avoid obstacles.
Combining radial velocities from on-chip radar with \acs{MEMS} inertial sensing has proven to provide robust, lightweight, and consistent state estimation, even in visually or geometrically degraded environments.
Statistical tests robustify these estimators against radar outliers.
However, available work with binary outlier filters lacks adaptability to various hardware setups and environments.
Other work has predominantly been tested in handheld static environments or automotive contexts.
This work introduces a robust \ac{BRIO} m-estimator for quadcopter flights in typical \acs{GNSS}-denied scenarios.
Extensive real-world closed-loop flights in cities and forests demonstrate robustness to moving objects and ghost targets, maintaining a consistent performance with \SI{0.5}{\percent} to \SI{3.2}{\percent} drift per distance traveled.
Benchmarks on public datasets validate the system's generalizability.
The code, dataset, and video are available at \url{https://github.com/ethz-asl/rio}.
\end{abstract}

\IEEEpeerreviewmaketitle

\section{Introduction}
Mobile robots are actively researched for disaster response.
Small, multirotor \acp{UAV} are ideal for exploration as they can traverse rubble and obstacles~\cite{roucek2021system}. 
In confined spaces \acp{UAV} need onboard positioning to correct \ac{IMU} drift.
Vision, lidar, and thermal navigation solutions have been proposed~\cite{qin2018vins,hudson2021heterogeneous,khattak2019robust}, but these are vulnerable to visual degradation, geometric ambiguity, or weak temperature gradients.

With the availability of automotive single-chip \ac{FMCW} radars~\cite{waldschmidt2021automotive}, \citet{doer2020ekf} demonstrated the first radar-inertial-stabilized quadcopter flights.
Due to its active sensing principle, radar is robust to visual degradation, such as darkness or fog. 
Further, radar's ability to measure linear displacements and \ac{IMU}'s ability to measure orientation changes make them an outstanding sensor combination to estimate the standard navigation state of position, velocity, and orientation.
This work considers the fusion of bearing Doppler radar detections with \ac{IMU} measurements.

\begin{figure}
    \begin{subfigure}{0.49\columnwidth}
        \centering
        \includegraphics[width=\textwidth]{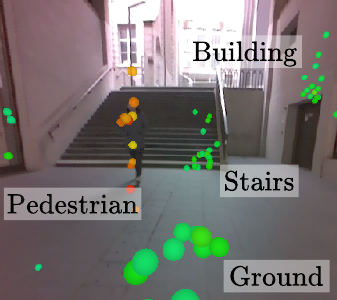}
        \caption{Urban environment hovering.}  
        \label{fig:dynamic_scene}
    \end{subfigure}
    \begin{subfigure}{0.49\columnwidth}
        \centering
        \includegraphics[width=\textwidth]{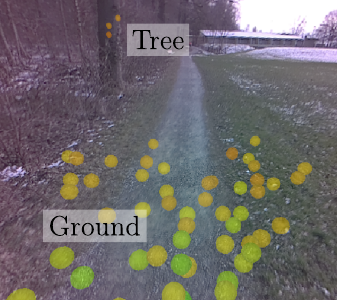}
        \caption{Forest path forward flight.}  
        \label{fig:forest_path}
    \end{subfigure}
    \caption{Onboard view from our radar-stabilized quadcopter. The points show \SI{1}{\second} accumulated, overlayed radar detections colored by their radial velocity (green: zero, orange: towards the radar, blue: away from the radar).}
    \label{fig:introduction}
    \vspace{-6mm}
\end{figure}

A core assumption in our work and others is that all radar detections are static.
However, moving objects and singular ghost targets introduced by multipath propagations or electromagnetic noise often violate this assumption.
\reffig{fig:dynamic_scene} shows that these outliers have a radial velocity inconsistent with the robot movement, and thus, statistical outlier tests robustify \iac{RIO} estimator.
Our work shows robust optimization sufficiently suppresses noise across realistic environments.
In contrast to binary outlier rejection, e.g., \acs{RANSAC}, m-estimation requires only a single tuning parameter.

A second necessity for estimating body motion is at least three radar detections with linearly independent \ac{DoA}.
Our experiments show that sufficient independent detections are generally given across human-made and natural environments.
The estimation is always stable, even in scenes where the bearing angle distribution is ill-conditioned, such as the dense ground reflections in \reffig{fig:forest_path}.

Finally, the \ac{DoA} may be biased, causing drift, especially in the gravity-affected elevation direction.
Differential barometry efficiently mitigates vertical drift.
Robust optimization also helps to reject noise induced by propeller downwash.

This work presents an experimental validation of \ac{BRIO} for quadcopter navigation in urban and natural environments.
It derives a factor-graph m-estimator with robust bearing Doppler and differential pressure factors.
Extensive experiments show estimator robustness against common outliers, such as cyclists, pedestrians, streetcars, multipath detections, and aerodynamic effects.
The estimated trajectory shows low drift and sufficient smoothness for multirotor feedback control.
Summarized, our contributions are
\begin{itemize}
    \item \Iac{BRIO} m-estimator with robust bearing Doppler, robust differential barometry, and zero-velocity factors.
    \item Experimental \rev{tuning}, robustness and performance analysis through radar-stabilized quadcopter flights in realistic natural and human-made environments.
    \rev{\item Demonstrating generalizability and state-of-the-art performance on a public dataset benchmark.}
    \item Open-source estimator, sensor drivers, and dataset.
\end{itemize}

\section{Related Work}
\citet{yokoo2009indoor} introduced the idea of pose estimation from sparse \ac{FMCW} radar bearing Doppler and \ac{IMU} measurements.
Consequently, \citet{kellner2013instantaneous} showed that cost-efficient, single-chip \ac{FMCW} radar delivers ego-motion estimation for autonomous driving.
They showed that two linearly independent radial velocity measurements are sufficient to compute a car's linear velocity.

\citet{doer2020ekf} generalized this concept to three-dimensional navigation.
They introduced a navigation filter combining \ac{IMU}  and barometry with linear velocities computed from at least three radial velocities and least squares.
However, their binary radar outlier tests require $16$ tuning parameters, complication setup adaptation.
Their barometry neglects aerodynamic disturbances and bias observability.

Their follow-up work uses multiple radars and introduces yaw estimation through a Manhatten world assumption~\cite{DoerJGN2022}.
Their results with additional sensors are mixed, and the assumption is unsuited for natural environments.
Instead, our results show sufficient yaw accuracy can be achieved on a consumer-grade \ac{IMU} using turn-on-bias calibration.

\citet{huang2023multi} examine how radar resolution affects state estimation accuracy.
A robust optimization computes the linear velocity from multiple radar detections and extracts the covariance.
A sliding window estimator fuses it with \ac{IMU} data.
Our experiments show that a simpler single radar-baro setup provides competitive performance and smooth state estimation for aerial robots in real-world environments.

\citet{michalczyk2022tightly} combine instantaneous radar velocity updates with radar landmark tracking in a filter framework.
The follow-up work achieves a drift between \SIrange[]{0.17}{1.62}{\percent}~\cite{michalczyk2023multi}.
However, feature tracking and mapping add complexity and are a source of error~\cite{kubelka2023we}.
In particular, generalizing feature association is challenging due to the fluctuating nature of radar reflections~\cite[pp.~106-108]{jankiraman2018fmcw}.
\citet{michalczyk2023multi} only present results in a small lab environment with artificially placed corner reflectors.
We limit tracking to static scenes to detect standstill.
Thus, we avoid restrictive radar target assumptions.
Our approach has similar accuracy while showing broader applicability.

\citet{kramer2021radar} present a fixed-lag smoother, similar to a total least squares problem.
They co-estimate robot velocity, orientation, radar detection \ac{DoA}, and Doppler velocity.
However, to stabilize the optimizer, they require \acs{MLESAC} prefiltering on top of robust loss.
Our approach considers radar reflections fixed, has fewer optimization variables, and does not require prefiltering.

\citet{kramer2020radar} work is closest to our approach.
They directly fuse individual bearing Doppler measurements in a robust optimization using Cauchy loss to reject radar outliers.
However, our work goes far beyond their work.
We compare different loss functions and demonstrate robust barometer fusion, closed-loop quadrotor control, robustness to dynamic objects, position estimation, turn-on-bias calibration, and zero-velocity tracking. Our work is available as open-source.

Finally, multiple radar state estimation approaches tailored for ground robots exist\rev{~\cite{harlow2023new}}.
However, most, such as full SLAM pipelines or large spinning radars on autonomous cars, are unsuited for quadcopter control, which requires lightweight velocity estimation and not a dense map.

\section{Notation}
Capital letters denote right-handed coordinate frames.
The origin of a specific coordinate frame is denoted by the same letter.
Matrices and vectors are bold.
$_A \mathbf{t} _{BC}$ is the vector from source $B$ to target $C$, expressed in coordinate frame $A$.
$_A\mathbf{e}_B\in S^2$ is the unit vector on the 2-sphere pointing from the origin of coordinate frame $A$ towards location $B$.

The direction cosine matrix ${\mathbf{R}_{AB}\in SO(3)}$ represent a rotation and maps vector ${_B \mathbf{t}}_{AB}$ from frame $B$ to frame $A$.
\begin{align}
\rev{{_A \mathbf{t}}_{AB}} = \mathbf{R}_{AB} \cdot {_B \mathbf{t}}_{AB}
\end{align}
Consequently, a rotation $\mathbf{R}_{AC}$ can be expressed by
\begin{align}
\mathbf{R}_{AC} = \mathbf{R}_{AB}\mathbf{R}_{BC}.
\end{align}
A rigid transformation $\mathbf{T}_{AB}\in SE(3)$ is a composition of the rotation $\mathbf{R}_{AB}\in SO(3)$ and translation $_A\mathbf{t}_{AB} \in \mathbb{R}^3$.

\section{Radar-inertial State Estimation}
\begin{figure}
    \centering
    \includegraphics[width=\columnwidth]{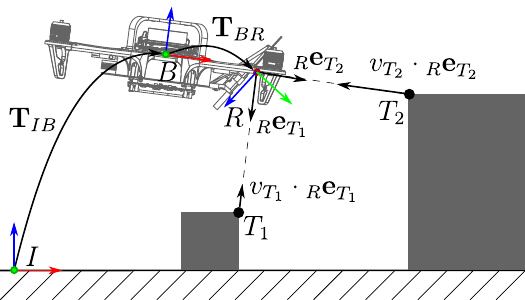}
    \caption{Quadcopter side view with coordinate frames and two radar detections.}
    \label{fig:coordinate_frames}
    \vspace{-4mm}
\end{figure}
The system is described by the three coordinate frames, denoted by capital letters $\gls{frame_inertial}$, $\gls{frame_body}$, and $\gls{frame_radar}$ in \reffig{fig:coordinate_frames}.
The unit vectors $\gls{frame_inertial_x}$, $\gls{frame_inertial_y}$, $\gls{frame_inertial_z}$ define the inertial frame $I$, where $\gls{frame_inertial_x}$ and $\gls{frame_inertial_y}$ span a plane tangential to the earth ellipsoid and $\gls{frame_inertial_z}$ points up.
$B$ denotes the body frame which coincides with the \ac{IMU} axes with $\gls{frame_body_x}$ forward, $\gls{frame_body_y}$ left, and $\gls{frame_body_z}$ up.
$R$ is the radar frame, where $\gls{frame_radar_x}$ points at positive azimuth direction, $\gls{frame_radar_y}$ at boresight, and $\gls{frame_radar_z}$ at positive elevation direction.
The coordinate frames are linked by rigid transformation $\gls{robot_pose}$ and $\gls{radar_calibration}$.
$\gls{robot_pose}$ is the online estimated, time-varying robot pose.
$\gls{radar_calibration}$ is the extrinsic calibration between \ac{IMU} and radar and obtained from \acs{CAD}.
\subsection{System State}
The \ac{IMU} pose $\gls{robot_pose}[^i] \in SE(3)$, composed of orientation $\gls{robot_orientation}[^i] \in SO(3)$ and translation $\gls{robot_translation}[^i] \in \mathbb{R}^3$, the body velocity $\gls{robot_linear_velocity}[^i] \in \mathbb{R}^3$, the gyroscope biases $\gls{bias_gyro}[^i] \in \mathbb{R}^3 $, and the accelerometer biases $\gls{bias_acc}[^i] \in \mathbb{R}^3$ form the $15$-dimensional state $\gls{state_vector}$ at time $\gls{index_time}$.
Our framework also tracks the location of zero velocity radar detections $\gls{track}^j$ with track index $\gls{index_track}$, where $\gls{track_position}[^{i,\gls{index_detection}}]$ is the estimated position of the $m$-th radar detection observed at time $i$, tracked over multiple frames $\gls{set_track} \subseteq \mathcal{K}_k$.
\begin{align}
    \gls{state_vector}^i &= \left[ \gls{robot_orientation}[^i],~ _I \mathbf{t} _{IB}^i,~ {_I \mathbf{v}} _{IB}^i,~ _B \mathbf{b}_g^i,~ _B \mathbf{b}_a^i \right]^T & \gls{track}[^j] &=  \left[ \gls{track_position}[^{i,m}] \right]^T
\end{align}

The estimator creates a new state with each radar frame.
\gls{set_state} is the set of all estimated states with $\gls{set_radar}$ the set of all radar frames up to time $k$.
Analogously, $\gls{set_track_estimates}$ is the set of all estimated zero velocity detection positions with $\gls{set_tracks}$ the set of all tracks up to time $k$.
\begin{align}
    \gls{set_state}  &= \{ \mathbf{x}^i \}_{i \in \mathcal{K}_k} & \mathcal{T}_k &= \{ \gls{track}[^j] \}_{j \in \mathcal{L}_k}
\end{align}
 
\subsection{Measurements}
The inputs to our system are bearing Doppler radar detections, \ac{IMU} measurements, and optionally barometric pressure.
We denote $\gls{set_detections}[^i]$ as the set of all detections at time $i$.
$\gls{set_imu}[^{i,i+1}]$ is the set of \ac{IMU} measurements between radar frame $i$ and $i+1$.
$\gls{set_baro}^i$ is the pressure measurement closest to time $i$.  
The measurement set up to time $\gls{index_current_time}$ is
\begin{align}
    \gls{set_meas} &= \{ \gls{set_detections}[^i], \mathcal{I}^{i,i+1}, \mathcal{B}^i \}_{(i,i+1) \in \mathcal{K}_k}.
\end{align}

\subsection{Optimization Criterion}
\begin{figure}
    \centering
    \includegraphics[page=2,width=\columnwidth]{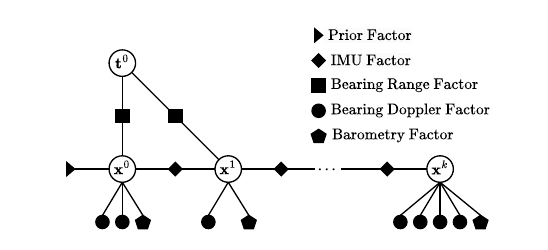}
    \caption{Factor graph representation with two, one, and four Doppler measurements and a zero-velocity landmark observed from the first and second state.} 
    \label{fig:factor_graph_online}      
    \vspace{-4mm}
\end{figure}
The online estimator evaluates the \ac{MAP} of states $\mathcal{X}_k$ and zero-velocity tracks $\mathcal{T}_k$ given all available observations $\mathcal{Z}_k$.
\begin{align}
\gls{set_state_map} &= \arg \max_{\mathcal{X}_k, \mathcal{T}_k} ~ p(\mathcal{X}_k, \mathcal{T}_k \vert \mathcal{Z}_k )
\end{align}
The factor graph in \reffig{fig:factor_graph_online} represents the unnormalized posterior~\cite{dellaert2017factor}.
\begin{align}
\begin{split}
\mathcal{X}_k^{\mathrm{MAP}} =& \arg \min_{\mathcal{X}_k, \mathcal{T}_k} \underbrace{\lVert \gls{residual_prior}[^0] \rVert_{\gls{cov_prior}}^2}_\text{(prior)} 
+ \sum_{(i,i+1) \in \mathcal{K}_k} \underbrace{ \lVert \gls{residual_imu}[^{i,i+1}] \rVert_{\gls{cov_imu}}^2}_\text{(IMU)} \\ 
&+ \sum_{j \in \mathcal{L}_k} \sum_{i \in \mathcal{K}_j} \underbrace{ \lVert \gls{residual_zero_velocity}[^{i,j}] \rVert_{\mathbf{\Sigma_T}}^2 }_ \text{(zero-velocity track)} \\
&+ \sum_{i \in \mathcal{K}_k} \sum_{m \in \gls{set_detections}[^i]} \underbrace{ \gls{robust_loss_function} \left( \frac{\lVert \gls{residual_doppler}[^{i,m}]\rVert}{\gls{std_doppler}} \right) \lVert \gls{residual_doppler}[^{i,m}] \rVert^2_{\gls{cov_doppler}}}_ \text{(robust bearing Doppler)} \\ 
&+ \sum_{i \in \mathcal{K}_k} \underbrace{ \gls{robust_loss_function} \left( \frac{\lVert \gls{residual_baro}[^i]\rVert}{\gls{std_baro}} \right) \lVert \gls{residual_baro}[^i] \rVert_{\gls{cov_baro}}^2}_ \text{(robust barometry)},
\end{split}
\label{eq:map_online}
\end{align} 
where $\gls{residual_prior}$ are prior, $\gls{residual_imu}$ \ac{IMU}, $\gls{residual_zero_velocity}$ zero-velocity track, $\gls{residual_doppler}$ bearing Doppler, and $\gls{residual_baro}$ barometry factor residual errors.
$\gls{cov_prior}$, $\gls{cov_imu}$, $\gls{cov_track}$, $\gls{cov_doppler}$, $\gls{cov_baro}$, $\gls{std_doppler}$, and $\gls{std_baro}$ are the corresponding measurement covariances respectively standard deviations. 
$\lVert \epsilon \rVert _\mathbf{\Sigma}^2 = \epsilon^T \mathbf{\Sigma}^{-1} \epsilon$ is the squared Mahalanobis distance with covariance $\mathbf{\Sigma}$.
$\gls{robust_loss_function}$ is a robust loss function specified in Sections \ref{sec:dopplerfactor} and \ref{sec:barofactor}.


\subsection{Robust bearing Doppler Factor}
\label{sec:dopplerfactor}
We consider \ac{FMCW} radar with a standard detection processing chain.
Upon receiving an asynchronous triggering signal, the \ac{MIMO} radar emits a predefined \ac{FMCW} chirp sequence.
The returning analog signal is mixed with the transmitting signal, converted to digital, windowed, range, and Doppler processed, \ac{CFAR} filtered, and \ac{DoA} processed.
The \ac{CFAR} detection extracts strong reflections separated in the range-Doppler space. 
\ac{CFAR} detection is an efficient data reduction step as \ac{DoA} processing and data transfer to the host computer only need to be done on a few data points.
Also, the \ac{CFAR} effectively filters reflections from noise. 

The result is a sparse pointcloud with detections $\gls{set_detections}[^i]$ at time ${i\in\rev{\gls{set_radar}}}$.
Detection ${\gls{radar_detection}[^{i,m}] \in \gls{set_detections}[^i]}$ contains \acl{DoA} ${\gls{radar_detection_bearing}[^{i,m}] = \frac{\gls{radar_detection_position}[^{i,m}]}{\lVert \gls{radar_detection_position}[^{i,m}] \rVert}}$, Doppler velocity ${\gls{radar_detection_doppler}[^{i,m}]}$, range ${r_{T_m}^i = \lVert \gls{radar_detection_position}[^{i,m}] \rVert}$, \ac{SNR} ${\gls{radar_detection_snr}[^{i,m}]}$, and noise $\gls{radar_detection_noise}[^{i,m}]$.
\reffig{fig:introduction} shows these detections mostly occur on geometric edges and strong changes in relative permittivity, e.g., from air to metal on the window frame.
We use the \ac{DoA} and Doppler velocity to constrain the body velocity into the direction of the detection as indicated in \reffig{fig:coordinate_frames}.

The bearing Doppler residual $\gls{residual_doppler}[^{i,m}]$ is the difference between the estimated radar linear velocity ${\gls{vel_radar}[^i]}$ projected onto the detection direction and the measured Doppler.
It depends on the state variables $\gls{robot_orientation}[^i]$, $\gls{robot_linear_velocity}[^i]$, and $\gls{bias_gyro}[^i]$ , radar measurements $\gls{radar_detection_bearing}[^{i,m}]$ and $\gls{radar_detection_doppler}[^{i,m}]$, and angular velocity $\gls{robot_angular_velocity}[^i]$. 
\begin{align}
    \gls{residual_doppler}[^{i,m}] &= -\left({\gls{vel_radar}[^{i}]}\right)^T \cdot \gls{radar_detection_bearing}[^{i,m}] - \gls{radar_detection_doppler}[^{i,m}] \\
    &= - \left(\left( \gls{robot_orientation}[^i] \gls{radar_calibration_rotation} \right)^{T} \cdot  \gls{vel_radar_inertial}[^{i}]\right)^T \cdot \gls{radar_detection_bearing}[^{i,m}]  - \gls{radar_detection_doppler}[^{i,m}], 
\end{align}
with the radar velocity $\gls{vel_radar_inertial}[^{i}]$ derived from the body velocity.
\begin{align}
    \gls{vel_radar_inertial}[^{i}] &=  \gls{robot_linear_velocity}[^i] + \gls{robot_orientation}[^i] \left( \left(\gls{robot_angular_velocity}[^i] - \gls{bias_gyro}[^i] \right) \times  \gls{radar_calibration_translation}  \right)
    \label{eq:radarextrinsics}
\end{align}

\begin{figure}
    \centering
    \begin{subfigure}{\columnwidth}
        \centering
        \includegraphics[width=\columnwidth]{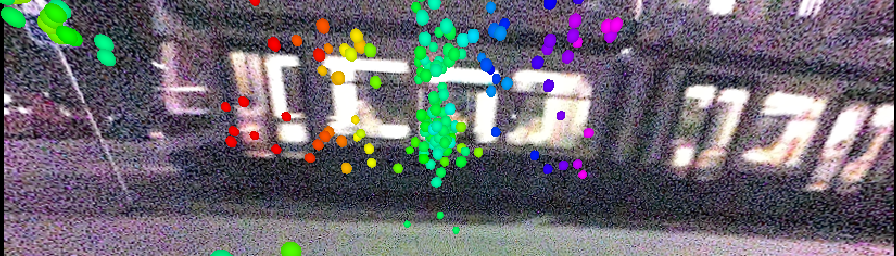}
        \caption{Quadcopter onboard view. Streetcar passing from left to right.}
        \label{fig:streetcar}
    \end{subfigure}
    \begin{subfigure}{\columnwidth}
        \centering
        \includegraphics{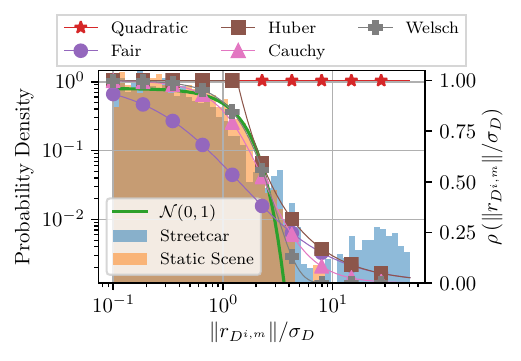}
        \caption{Normalized absolute bearing Doppler pre-fit residuals across one optimization window with and without moving object in the radar field of view. $\gls{std_doppler}=0.05$, and optimizer using Welsch loss. Robust loss functions tuned to \SI{95}{\percent} asymptotic efficiency~\cite{rey2012introduction}.}
        \label{fig:residuals_doppler}      
    \end{subfigure}
    \begin{subfigure}{\columnwidth}
        \centering
        \includegraphics{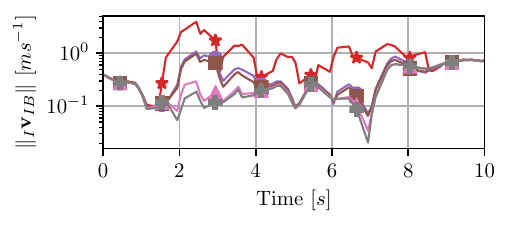}
        \caption{Estimated body velocity while hovering in front of the streetcar.}
    \label{fig:robust_hovering}      
    \end{subfigure}
    \caption{Robust bearing Doppler loss function analysis based on streetcar passing hovering quadcopter. Radar detections accumulated over \SI{10}{\second} optimization window.}
    \vspace{-4mm}
\end{figure}

In a static scene, the bearing Doppler residual noise is normally distributed with standard deviation $\gls{std_doppler}$ as shown in \reffig{fig:residuals_doppler}.
However, the radar occasionally detects ghost targets or moving objects, creating long-tailed residual noise.
For example, the streetcar in \reffig{fig:streetcar} generates approaching (orange/red) and receding (blue/purple) radial velocity outliers.
\reffig{fig:residuals_doppler} shows the corresponding bearing Doppler residual error distribution with a visible outlier set located between \SIrange[]{10}{50}{} standard deviations.
The following evaluated loss functions~\cite{rey2012introduction} suppress these outliers to varying degrees.

Our experiments with moving objects have shown that a loss function should be selected where the weighting function $\gls{robust_loss_function}$ approaches zero.
\reffig{fig:robust_hovering} shows the velocity estimate during hover while the streetcar passes.
The estimate should be close to zero.
A pure quadratic loss causes significant non-zero velocity estimates.
Also, Fair and Huber's losses are insufficient to suppress the streetcar detections.
Even Cauchy loss will still predict some movement in the presence of outliers.
Welsch loss shows the steadiest hover estimate and is selected for the estimation problem \refequ{eq:map_online}.

One beauty of using only robust loss for outlier rejection compared to additional statistical tests, such as \ac{RANSAC}, is that only $\gls{std_doppler}$ requires tuning to normalize the residuals \refequ{eq:map_online}.
We keep $\gls{std_doppler}$ constant throughout all experiments.
We did not observe instabilities in the optimization with any of the evaluated $\gls{robust_loss_function}$ functions.

\subsection{Robust Barometric Factor}
\label{sec:barofactor}
In principle, the radar factor is sufficient to observe the body velocity and \rev{integrated} position if three linearly independent radial velocities are measured.
However, systematic bias in these measurements, paired with errors in the extrinsic calibration, gravity constant, and \ac{IMU} biases, cause altitude estimation drift.
To compensate for errors in height, we measure differential barometry.
The barometer residual $\gls{residual_baro}[^i]$ is the difference between estimated height, offset by bias $\gls{baro_meas}[^0]$, and measured height $\gls{baro_meas}[^i]$.
\begin{align}
    \gls{residual_baro}[^i] &= \gls{frame_inertial_z}^T \cdot \gls{robot_translation}[^i]  + \gls{baro_meas}[^0] - \gls{baro_meas}[^i],
\end{align}
with $\gls{baro_meas}[^i]$ a function of the pressure $\gls{baro_pressure_meas}[^i]$ at time $i$, derived from an earth atmosphere model~\cite{model12nasa}.
\begin{align}
    \gls{baro_meas}[^i] &= \frac{288.08 \left(\frac{\gls{baro_pressure_meas}[^i]}{101290}\right)^\frac{1}{5.256} - 273.1 - 15.04}{-0.00649}
\end{align}

Barometric measurements are biased by ambient temperature changes~\cite{parviainen2008differential}.
Unlike \citet{doer2020ekf}, who co-estimate the pressure bias $\gls{baro_meas}[^0]$~, we set it constant using the first barometer measurement.
We noticed that $\gls{baro_meas}[^0]$ is unobservable because the ambient pressure drift due to weather changes is much smaller than the \ac{RIO} height estimation drift.

\begin{figure}
    \centering
    \begin{subfigure}{\columnwidth}
        \centering
        \includegraphics[width=\columnwidth]{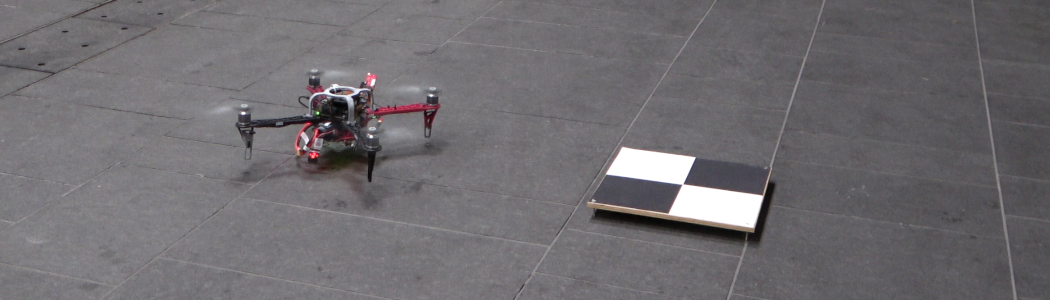}
        \caption{Photo of quadcopter landing next to take off location.}
        \label{fig:robust_landing}
    \end{subfigure}
    \begin{subfigure}{\columnwidth}
        \centering
        \includegraphics{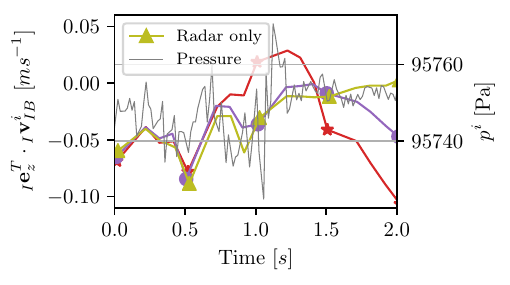}
        \caption{Vertical velocity estimate during landing at \SI{1}{\second} with quadratic and Fair loss. Pressure showing ground effect.}
    \label{fig:velocity_jump}      
    \end{subfigure}
    \begin{subfigure}{\columnwidth}
        \centering
        \includegraphics{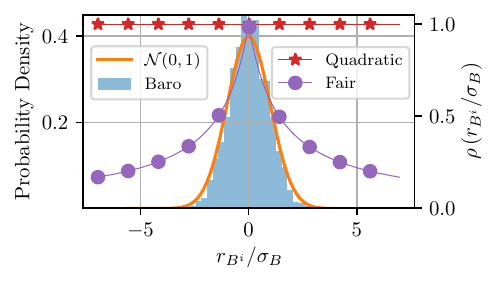}
        \caption{Normalized barometer pre-fit residual distribution over complete flight with $\gls{std_baro}=0.2$, and Fair barometer residual loss. Robust loss functions tuned to \SI{95}{\percent} asymptotic efficiency~\cite{rey2012introduction}}
    \label{fig:residuals_baro}      
    \end{subfigure}
    \caption{Barometer noise and robust loss analysis.}
    \vspace{-4mm}
\end{figure}

However, our estimator is robust to sudden disturbances, such as ground effects during touch-down (\reffig{fig:robust_landing}).
Without suppression, this noise can falsely deviate the vertical velocity estimate, as seen in red in \reffig{fig:velocity_jump}.
The Fair robust loss function ensures a smooth estimate.
Apart from these few aerodynamic occasions, the barometric residuals are normally distributed as shown in \reffig{fig:residuals_baro}.

\subsection{Zero-Velocity Detection Tracking}
\label{sec:zerovelocity}
Before take-off and after landing, all static reflections have zero Doppler velocity and cannot be separated in Doppler space by the \ac{CFAR} detector.
Thus, the radar point cloud often consists of only one or two targets, representing the strongest reflections that can be separated by range.
This is insufficient to compute the linear body velocity, so the state estimate will drift on a sphere or circle.
Since the environment and robot are static, consecutive reflections are identical within the sensor resolution.
We track these targets, making the full state observable at zero velocity.

Let $\gls{radar_detection}[^{i,m}] \in \gls{set_detections}[^i]$ be detection $m$ at time $i$ and $\gls{radar_detection}[^{\underline{i},\underline{m}}] \in \gls{set_detections}[^{\underline{i}}]$ be detection $\underline{m}$ at time $\underline{i}$.
Tracks are discarded after being unobserved for a fixed amount of radar frames.
Targets are associated if they have identical measurements.
\begin{align}
\begin{split}
 &\left( {\gls{radar_detection_position}[^{i,m}]} = {\gls{radar_detection_position}[^{\underline{i},\underline{m}}]} \right) \wedge 
 \left( {\gls{radar_detection_doppler}[^{i,m}]} = {\gls{radar_detection_doppler}[^{\underline{i},\underline{m}}]} = 0 \right) \wedge \\
 &\left( {\gls{radar_detection_snr}[^{i,m}] = \gls{radar_detection_snr}[^{\underline{i},\underline{m}}]} \right) \wedge 
 \left( {\gls{radar_detection_noise}[^{i,m}] = \gls{radar_detection_noise}[^{\underline{i},\underline{m}}]} \right) \\
& \Rightarrow  \gls{radar_detection}[^{i,m}] = \gls{radar_detection}[^{\underline{i},\underline{m}}]
\end{split}
\end{align}

The optimizer estimates the position of each track in the inertial frame.
The residual, when represented in Cartesian coordinates, is the difference between the estimated detection position $\gls{track}[^j]$ and the measured detection position $\gls{radar_detection_position}[^{i,m}]$.
The estimated detection location is transformed from the inertial to the radar frame.
\begin{align}
    \gls{residual_zero_velocity}[^{i,j}] &= \left( \gls{robot_orientation}[^i] \gls{radar_calibration_rotation} \right)^{T} \left( {\gls{track}[^j]} -  \gls{robot_translation}[^i] - \gls{robot_orientation}[^i] \cdot \gls{radar_calibration_translation} \right) -  {\gls{radar_detection_position}[^{i,m}]}
\end{align}

\subsection{IMU Factor}
Integrating high-rate \ac{IMU} measurements as individual factors quickly increases variables in the optimization, making it computationally intractable.
\citet{forster2015imu} presented a method to preintegrate a set of \ac{IMU} measurements into a single factor.
The method successively computes the relative change in orientation, velocity, position.
The combined \ac{IMU} residual between two states is
\begin{align}
    \gls{residual_imu}[^{i,i+1}] = \left[ \mathbf{r}_{\Delta\mathbf{R}_{IB}^{i,i+1}}, \mathbf{r}_{\Delta{\mathbf{t}}_{IB}^{i,i+1}}, \mathbf{r}_{\Delta{\mathbf{v}}_{IB}^{i,i+1}}, \mathbf{r}_{\Delta{\mathbf{b}_{g}^{i,i+1}}}, \mathbf{r}_{\Delta{\mathbf{b}_{a}^{i,i+1}}} \right].
\end{align}

\subsection{Prior Factor}
\label{sec:prior}
\begin{figure}
    \centering
    \includegraphics[page=2]{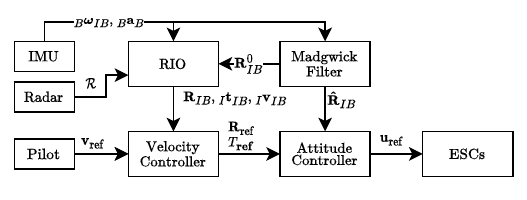}
    \caption{Quadcopter state estimation and controller diagram.}
    \label{fig:system}
    \vspace{-4mm}
\end{figure}

Proper optimizer initialization ensures a consistent state estimate from the start. 
The yaw bias and orientation require special handling.
Observing yaw bias is difficult due to the small lever arm \gls{radar_calibration_translation} and relatively large radar noise \gls{std_doppler}~\cite{DoerJGN2022}.
We calibrate the gyroscope bias by averaging the first messages and initializing it with high certainty.
A parallel Madgwick filter~\cite{ccny2024imu} aligns orientation with the attitude controller (\reffig{fig:system}).
All other values are initialized to zero.
\begin{align}
    \gls{residual_prior_orientation}[^0] &= Log\left( \left( \gls{robot_orientation}[^0] \right)^T \Tilde{\mathbf{R}}_{IB}^0\right) \\
    \gls{residual_prior_position}[^0] &= {_I\mathbf{t}_{IB}^0} - {_I\Tilde{\mathbf{t}}_{IB}^0}\\
    \gls{residual_prior_velocity}[^0] &= \gls{robot_linear_velocity}[^0] - {_I\Tilde{\mathbf{v}}_{IB}^0} \\
    \gls{residual_prior_bias_gyro}[^0] &= {_B\mathbf{b}}_g^{0} - {_B\Tilde{\mathbf{b}}}_g^{0}\\
    \gls{residual_prior_bias_acc}[^0] &= {_B\mathbf{b}}_a^{0} - {_B\Tilde{\mathbf{b}}}_a^{0},
\end{align}
with the logarithmic map transforming a rotation to its tangent space~\cite{sola2018micro} and tilde denoting initial values.
The combined prior factor is
\begin{align}
    \gls{residual_prior}[^0] = \left[ \gls{residual_prior_orientation}[^0], \gls{residual_prior_position}[^0], \gls{residual_prior_position}[^0], \gls{residual_prior_bias_gyro}[^0], \gls{residual_prior_bias_acc}[^0]
    \right].
\end{align}

\section{Experiments}
\begin{table*}
\centering
    \caption{Summary of experiments. The mode describes the experiment settings. H: Handheld, F: Flying}
    \resizebox{\textwidth}{!}{%
    \begin{tabular}{lccccccccccc}
        \hline
        Scenario & No. & Mode & Length & \makecell{Maximum \\ Velocity} & \makecell{Median \\ Detections} & \makecell{Moving \\ Objects} & \makecell{Final Position \\ Drift} & \makecell{\rev{XY-Position} \\ \rev{Drift}} & \makecell{Position Drift \\ per Distance} & \makecell{Final Yaw \\ Drift} & \makecell{Yaw Drift \\ \rev{per Distance}} \\
                   & & & [\si{\metre}] & [\si{\metre\per\second}]  &[\si{1}] &[\unit{1}] & [\si{m}] & [\si{m}] & [\si{\percent}]& [\si{\degree}] & [\si{\degree\per\metre}]\\
        \hline
        \arrayrulecolor{lightgray}
        Urban Night & 01 & \mbox{RIO-H} & 179.7 & 1.82 & 15 & 0 & 3.54 & 1.48 & 1.97 & -0.79 & -0.004\\ 
         &  & \mbox{BRIO-H} & 184.7 & 1.83 & 15 & 0 & 1.42 & 1.41 & 0.77 & -0.67 & -0.004\\ 
         \hline
         & 02 & \mbox{RIO-H} & 151.5 & 1.95 & 14 & 0 & 3.35 & 0.98 & 2.21 & -1.70 & -0.011\\ 
         &  & \mbox{BRIO-H} & 156.5 & 1.96 & 14 & 0 & 0.89 & 0.80 & 0.57 & -1.70 & -0.011\\ 
         \hline
         & 03 & \mbox{RIO-H} & 153.9 & 1.82 & 13 & 0 & 3.09 & 2.29 & 2.00 & 2.11 & 0.014\\ 
         &  & \mbox{BRIO-H} & 157.5 & 1.83 & 13 & 0 & 2.39 & 2.36 & 1.52 & 2.76 & 0.018\\ 
         \hline
         & 04 & \mbox{RIO-H} & 156.4 & 1.79 & 15 & 1 & 3.24 & 0.81 & 2.07 & 1.56 & 0.010\\ 
         &  & \mbox{BRIO-H} & 163.5 & 1.80 & 15 & 1 & 1.14 & 1.08 & 0.70 & 1.50 & 0.009\\ 
         \hline
         & 05 & \mbox{RIO-F} & 160.0 & 1.91 & 10 & 1 & 5.35 & 5.31 & 3.34 & 9.03 & 0.056 \\ 
         &  & \mbox{BRIO-F}  & 166.9 & 1.91 & 10 & 1 & 5.27 & 5.26 & 3.16 & 8.31 & 0.050\\ 
         \hline
         & 06 & \mbox{RIO-F} & 157.0 & 1.78 & 11 & 2& 4.79 & 4.57 & 3.05 & 8.94 & 0.057 \\ 
         &  & \mbox{BRIO-F} & 162.4 & 1.81 & 11 & 2 & 4.78 & 4.77 & 2.94 & 8.59 & 0.053 \\ 
         \hline
         & 07 & \mbox{RIO-F} & 185.4 & 1.62 & 11 & 0 & 4.36 & 4.09 & 2.35 & 8.98 & 0.048\\ 
         &  & \mbox{BRIO-F} & 194.9 & 1.62 & 11 & 0 & 4.41 & 4.27 & 2.26 & 8.69 & 0.045\\ 
         \hline
        Forest Path & 08 & \mbox{RIO-H} & 477.6 & 2.03 & 8 & 1 & 7.50 & 2.54 & 1.57 & 1.56 & 0.003 \\ 
         &  & \mbox{BRIO-H} & 480.3 & 2.07 & 8 & 1 & 2.77 & 2.52 & 0.58 & 1.15 & 0.002 \\ 
         \hline
         & 09 & \mbox{RIO-H} & 125.1 & 1.87 & 6 & 2 & 0.48 & 0.46 & 0.38 & 0.55 & 0.004 \\ 
         &  & \mbox{BRIO-H} & 126.4 & 1.90 & 6 & 2 & 0.77 & 0.50 & 0.61 & 0.57 & 0.005\\ 
         \hline
         & 10 & \mbox{RIO-H}  & 099.4 & 1.97 & 5 & 1 & 3.20 & 0.54 & 3.21 & 0.80 & 0.008 \\ 
         &  & \mbox{BRIO-H} & 101.5 & 2.00 & 5 & 1 & 0.51 & 0.46 & 0.50 & 0.94 & 0.009\\ 
         \hline
         & 11 & \mbox{RIO-F} & 487.4 & 1.98 & 7 & 0 & 17.19 & 15.00 & 3.53 & 12.41 & 0.025 \\ 
         &  & \mbox{BRIO-F} & 492.8 & 1.97 & 7 & 0 & 13.56 & 13.56 & 2.75 & 11.80 & 0.024\\ 
         \hline
         & 12 & \mbox{RIO-F} & 487.5 & 2.24 & 6 & 0 & 13.23 & 12.83 & 2.71 & 7.56 & 0.016 \\ 
         &  & \mbox{BRIO-F} & 490.4 & 2.25 & 6 & 0 & 11.77 & 11.37 & 2.40 & 6.58 & 0.013 \\ 
         \hline
        Flat Field & 13 & \mbox{RIO-F} & 198.9 & 2.54 & 6 & 0 & 4.55 & 3.50 & 2.29 & 5.63 & 0.028 \\ 
         &  & \mbox{BRIO-F} & 202.7 & 2.62 & 6 &  0 & 3.43 & 3.23 & 1.69 & 6.08 & 0.030  \\ 
         \hline
         & 14 & \mbox{RIO-F} & 295.2 & 2.08 & 9 & 3 & 7.15 & 6.99 & 2.42 & 8.57 & 0.029 \\ 
         &  & \mbox{BRIO-F} & 299.0 & 2.08 & 9 & 3 & 5.93 & 5.92 & 1.98 & 8.32 & 0.028 \\ 
         \hline
        Tree Slalom & 15 & \mbox{RIO-F} & 159.2 & 2.37 & 8 & 0 & 3.56 & 0.64 & 2.24 & 4.30 & 0.027 \\ 
         &  & \mbox{BRIO-F} & 161.9 & 2.37 & 8 & 0 & 0.74 & 0.59 & 0.46 & 4.0 & 0.029 \\ 
    \arrayrulecolor{black}
    \hline
    \end{tabular}}
    \vspace{-4mm}
    \label{tab:experiments}
\end{table*}

\label{sec:experiments}
We evaluate our approach with radar-stabilized quadcopter flights.
The pilot gives velocity setpoints via RC, and if no input is given, the drone hovers stably.
Controller, state estimation, and logging run onboard a Nvidia Jetson Orin NX, communicating with a BMI088 \ac{IMU}, a AWR1843AOPEVM radar, a BMP390 barometer, and ESCs through Linux user space drivers and ROS middleware (\reffig{fig:system}).
The optimization, implemented with GTSAM~\cite{gtsam}, runs in two threads using the iSAM2 solver~\cite{kaess2012isam2}.
An optimization thread solves the \ac{MAP} with a \SI{10}{\second} data window typically within \SI{40}{\milli\second} upon radar measurement arrival.
A navigation thread outputs the latest state prediction at \ac{IMU} rate to the controller~\cite{indelman2013information}.

The \ac{IMU} is tuned to capture the drone dynamics but digitally filter as much noise as possible.
The accelerometer has a range of \SI{6}{\g}, $4$-fold oversampling, and \SI{400}{\Hz} rate.
The synchronized gyroscope has a range of \SI{250}{\degree\per\second}, and a \SI{47}{\Hz} low-pass filter.
The radar maximizes velocity resolution
at \SI{0.04}{\metre\per\second}, with \SI{2.56}{\metre\per\second} maximum radial velocity and \SI{10.95}{\metre} maximum range.
The \ac{CFAR} detection threshold is \SI{15}{\decibel} for normally distributed residuals.
Radar frames are asynchronously triggered at \SI{8}{\Hz}.
The barometer runs asynchronously at \SI{50}{\Hz} with $8$-fold oversampling and a cumulative average filter with window size $3$.
Zero-order-hold interpolation synchronizes the three sensor timestamps.

In the following, \textit{RIO} refers to our optimization without barometer factor and \textit{BRIO} with it.
All flight experiments used \textit{RIO} controller feedback.
In addition, we collected data handheld.
An experiment starts and ends at the same location with an accuracy of approximately \SI{2}{cm} and \SI{2}{\degree} in the heading to compute the final drift.
\reftab{tab:experiments} summarizes all experiments, with numerical values obtained by rerunning the algorithm on collected raw sensor data.

\subsection{Exceptional Robustness}
Our main result is the outstanding robustness of radar navigation.
Unlike vision or lidar, radar measures velocities, avoiding feature tracking failures that cause state estimation divergence and crashes.
Seven handheld experiments (\SI{20}{\minute} total) and eight flights (\SI{35}{\minute} total) without failure demonstrate this robustness.
\textit{RIO} in the control loop shows sufficient latency and smoothness.

\begin{figure*}
    \begin{subfigure}{0.32\textwidth}
        \centering
        \includegraphics[width=\textwidth]{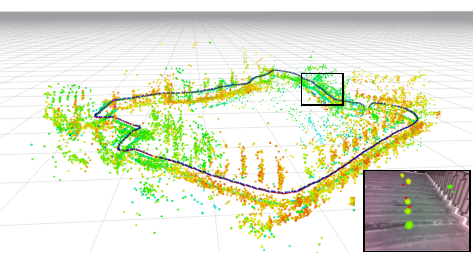}
        \caption{Urban Night 02 BRIO-H}
        \label{fig:rviz_urban_tilted}
    \end{subfigure}
    \hfill
    \begin{subfigure}{0.32\textwidth}
        \centering
        \includegraphics[width=\textwidth]{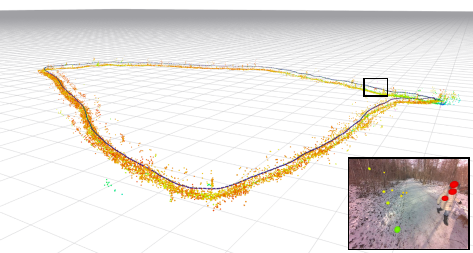}
        \caption{Forest Path 08 BRIO-H}
        \label{fig:rviz_forest_tilted}
    \end{subfigure}
    \hfill
    \begin{subfigure}{0.32\textwidth}
        \centering
        \includegraphics[width=\textwidth]{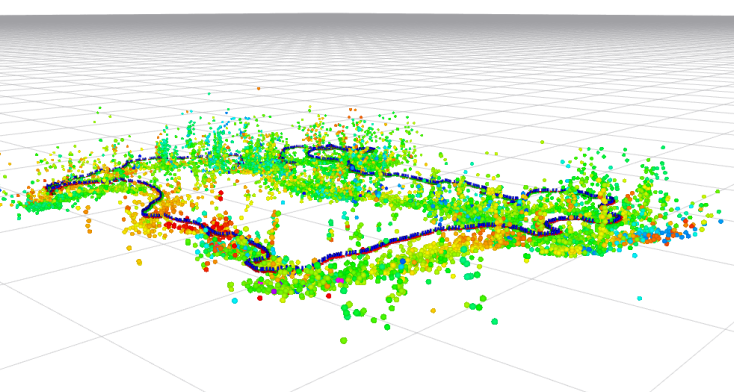}
        \caption{Tree Slalom 15 BRIO-F}
        \label{fig:rviz_slalom_tilted}
    \end{subfigure}
    \caption{Isometric view of accumulated point cloud and poses for selected experiments. Grid with \SI{10}{\metre} spacing and Doppler point coloring.
    Photos show three multipath detections on stairs and a pedestrian detection on the forest path (red points).
    }
    \vspace{-4mm}
    \label{fig:rviz}
\end{figure*}

The median number of detected points per scan drops in natural forests compared to urban environments.
The urban environment has more objects with high \ac{RCS}, such as the window frames in \reffig{fig:dynamic_scene}.
However, detections from trees or the ground are generally sufficient to estimate the body velocity.
\reffig{fig:rviz} shows accumulated point clouds for different environments.
\textit{Urban Night} (\reffig{fig:rviz_urban_tilted}) and \textit{Tree Slalom} (\reffig{fig:rviz_slalom_tilted}) have points distributed in elevation, while \textit{Forest Path} (\reffig{fig:rviz_forest_tilted}) is mostly flat with occasional tree trunks.
This demonstrates robustness in geometrically degenerate environments where \acl{LIO} can fail.
Additionally, \reffig{fig:rviz_urban_tilted} shows multipath detections on the stairs and \reffig{fig:rviz_forest_tilted} a pedestrian detection.
Both leave the estimate unaffected, highlighting the robustness of the selected Welsch bearing Doppler loss.

\subsection{Performance Overview}
\begin{figure*}
    \centering
    \begin{subfigure}{0.19\textwidth}
        \centering
        \includegraphics[width=\textwidth]{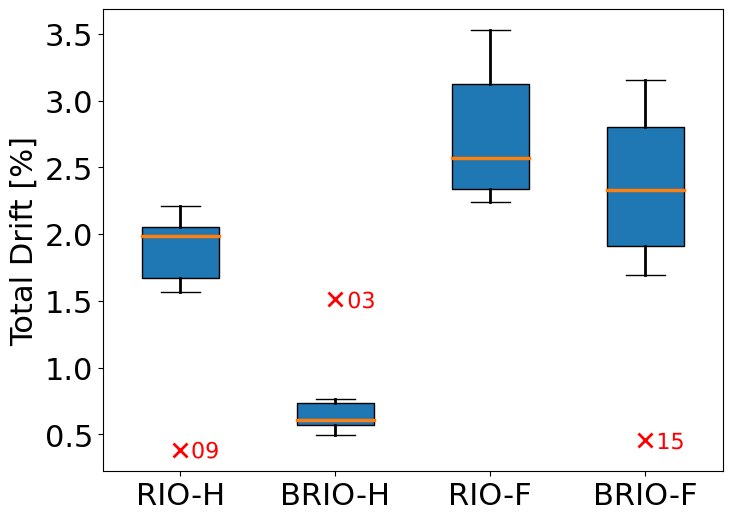}
        \caption{$\gls{frame_inertial_x}$, $\gls{frame_inertial_y}$, $\gls{frame_inertial_z}$ all.}
        \label{fig:total_drift_handheld_flight}        
    \end{subfigure}
    \hfill
    \begin{subfigure}{0.19\textwidth}
        \centering
        \includegraphics[width=\textwidth]{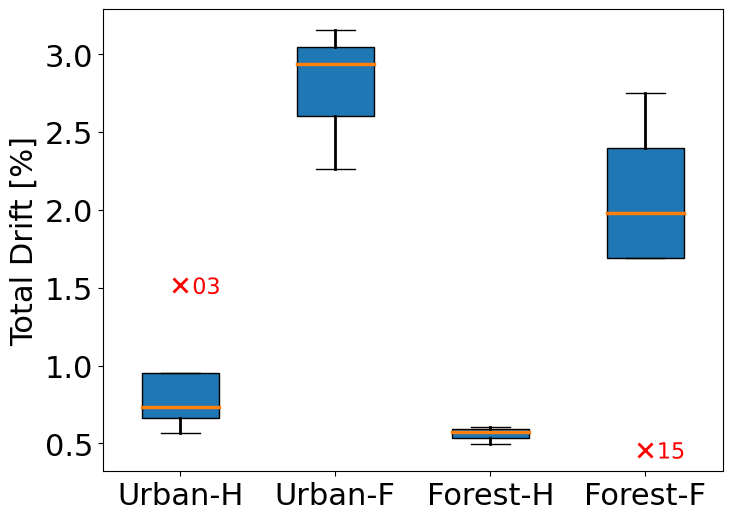}
        \caption{$\gls{frame_inertial_x}$, $\gls{frame_inertial_y}$, $\gls{frame_inertial_z}$ \textit{BRIO}.}
        \label{fig:total_drift_env_handheld_flight}
    \end{subfigure}
    \hfill
    \begin{subfigure}{0.19\textwidth}
        \centering
        \includegraphics[width=\textwidth]{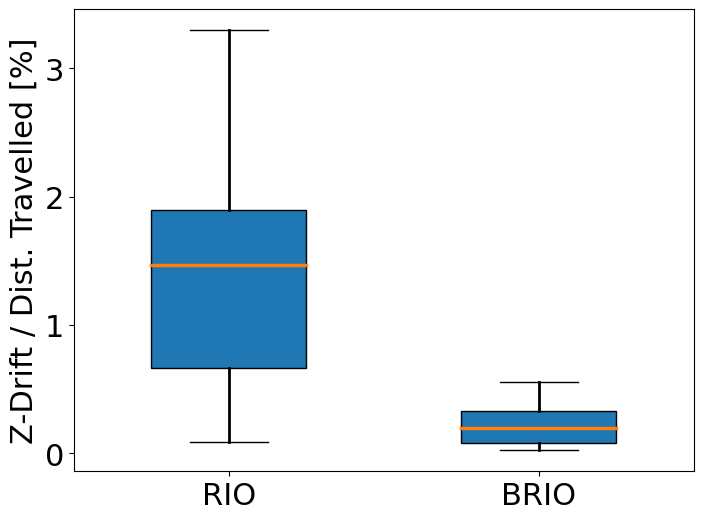}
        \caption{$\gls{frame_inertial_z}$ all.}
        \label{fig:z_drift_allexp}
    \end{subfigure}
    \hfill
    \begin{subfigure}{0.19\textwidth}
        \centering
        \includegraphics[width=\textwidth]{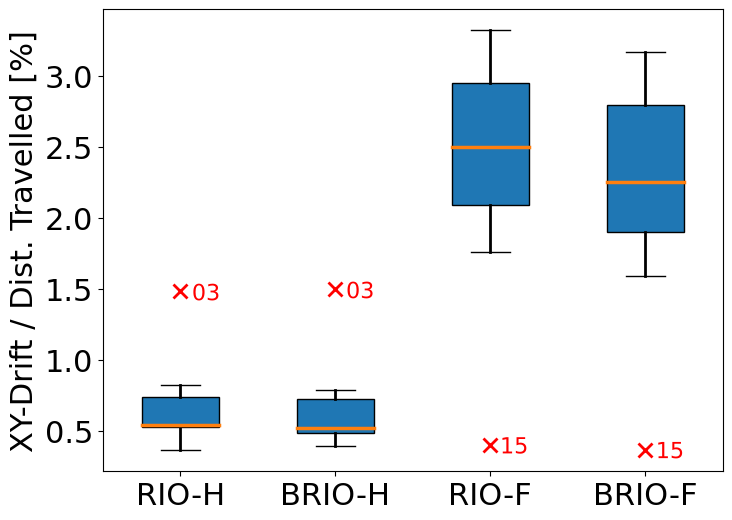}
        \caption{$\gls{frame_inertial_x}$, $\gls{frame_inertial_y}$ all.}
        \label{fig:xy_drift_handheld_flight}        
    \end{subfigure}
    \hfill
    \begin{subfigure}{0.19\textwidth}
        \centering
        \includegraphics[width=\textwidth]{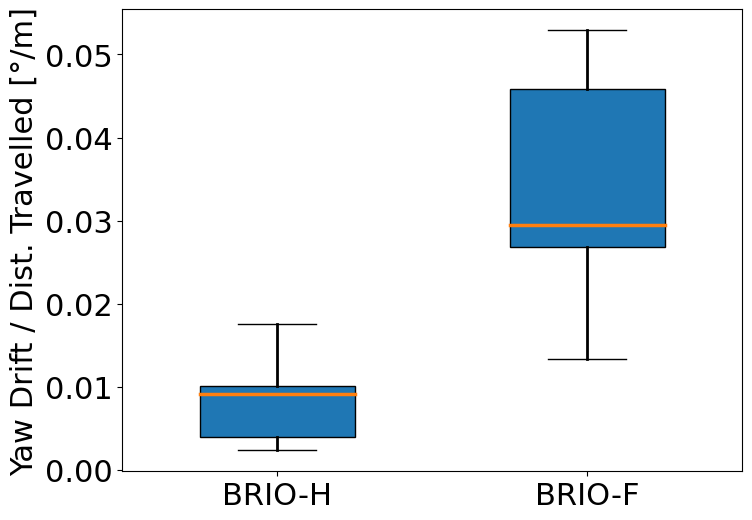}
        \caption{Heading drift all.}
        \label{fig:yaw_drift_handheld_flight}
    \end{subfigure}
    \caption{Start-to-end drift comparison.}
    \vspace{-4mm}
\end{figure*}

\reffig{fig:total_drift_handheld_flight} reports less than \SI{1}{\percent} start-to-end drift in \textit{BRIO} handheld experiments and \SIrange{2}{3}{\percent} in \textit{RIO} and \textit{BRIO} flights.
The state estimate performs consistently across environments (\reffig{fig:total_drift_env_handheld_flight}).
The best result, \SI{0.5}{\percent} drift over a \SI{162}{\metre} trajectory, was in the \textit{Tree Slalom 15 BRIO-F} experiment.
The results align with other state-of-the-art \ac{RIO} estimators~\cite{doer2020ekf,michalczyk2023multi}.
Well-tuned \ac{VIO} estimators report drift as low as \SI{0.3}{\percent}~\cite{qin2018vins} but are vulnerable to poor lighting conditions, as in \textit{Urban Night} (\reffig{fig:streetcar}).

Barometry significantly reduces vertical drift (\reffig{fig:z_drift_allexp}).
However, its influence on total xyz-drift (\reffig{fig:total_drift_handheld_flight}) depends on the horizontal drift magnitude.
Handheld experiments mostly drift vertically.
Thus the barometer cuts total drift from \SI{2}{\percent} to almost \SI{0.5}{\percent}.
Flights are dominated by horizontal drift (\reffig{fig:xy_drift_handheld_flight}).
Hence, total drift remains between \SIrange{2}{3}{\percent}.
\begin{figure*}
    \centering
    \begin{subfigure}{0.3\textwidth}
        \centering
        \includegraphics[width=\textwidth]{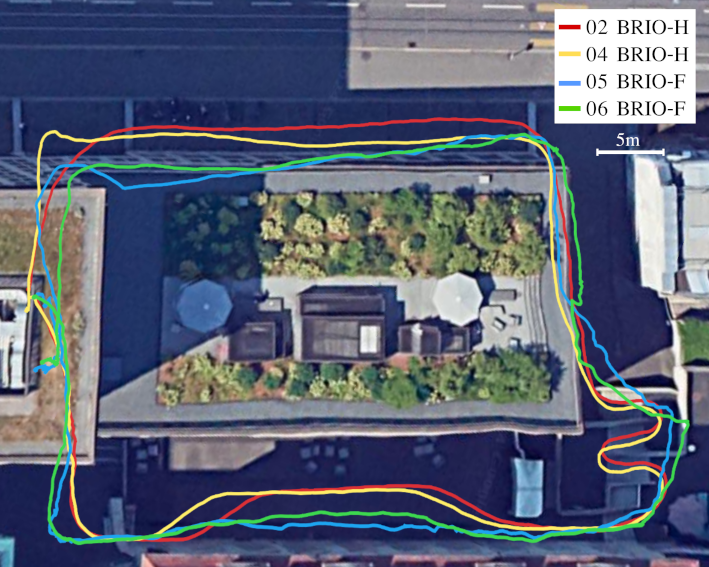}
        \caption{Urban Night}
        \label{fig:sat_urban}
    \end{subfigure}
    \hfill
    \begin{subfigure}{0.3\textwidth}
        \centering
        \includegraphics[width=\textwidth]{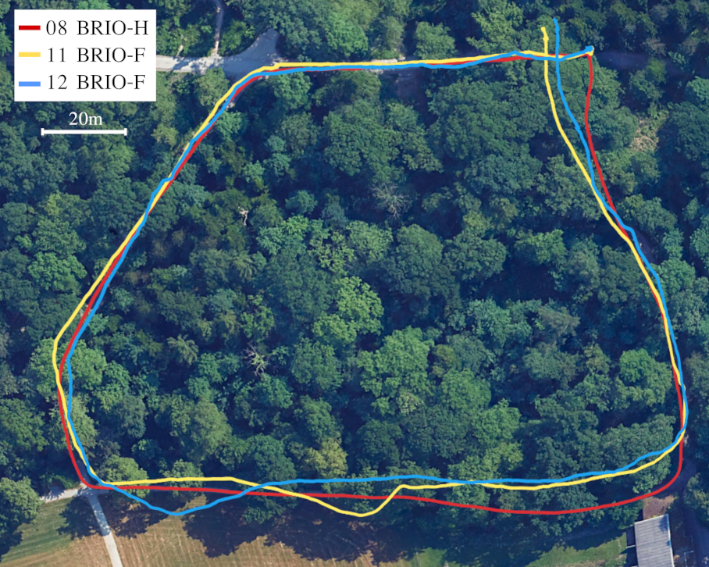}
        \caption{Forest Path.}
        \label{fig:sat_forest}
    \end{subfigure}
    \hfill
    \begin{subfigure}{0.3\textwidth}
        \centering
        \includegraphics[width=\textwidth]{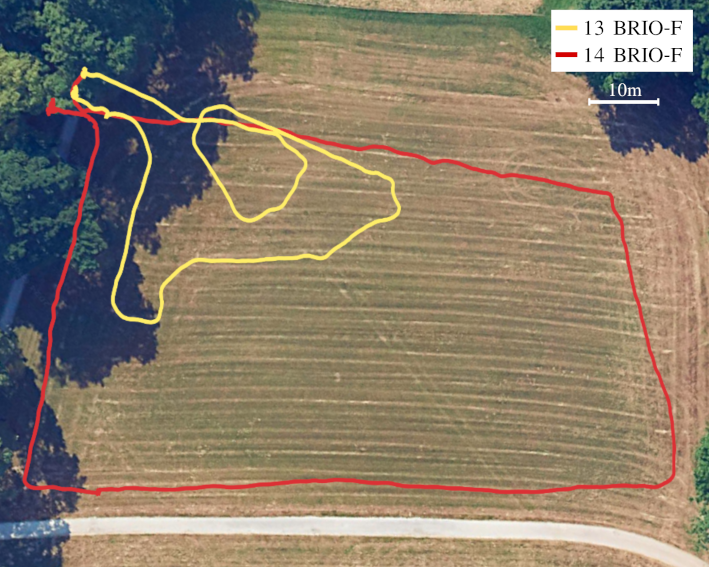}
        \caption{Flat Field.}
        \label{fig:sat_field}
    \end{subfigure}
    \caption{Satellite image trajectory overlay. Trajectories are manually aligned.}
    \vspace{-6mm}
    \label{fig:sat}
\end{figure*}
\subsection{Discussion: Vibrations Leading to Horizontal Flight Drift}
\reffig{fig:xy_drift_handheld_flight} shows that the horizontal drift increases from \SI{0.5}{\percent} to almost \SI{2.5}{\percent} when comparing handheld to flown experiments. 
We attribute this error to gyroscope scale error introduced by platform vibrations~\rev{\cite[p.~49]{braun2016high}}.
\reffig{fig:yaw_drift_handheld_flight} shows that there is a \rev{\SI{0.03}{\degree\per\metre}} yaw drift in flight versus \rev{\SI{0.01}{\degree\per\metre}} handheld.
The yaw drift is best seen in the trajectory overview in \reffig{fig:sat_forest}, where it causes the flown trajectory estimates to twist inwards \rev{on the four left-hand bends}, leading to horizontal drift.
Our quadcopter design does not dampen the low-cost \ac{IMU} likely causing vibration rectification~\cite{vibrationrect}.

The \rev{scale-error} possibly also explains the lowest drift in \textit{Tree Slalom 15 BRIO-F}.
\reffig{fig:rviz_slalom_tilted} shows the slalom flight is dynamic with many changes of directions, causing various vibration modes.
These may cause alternating yaw drift, potentially canceling each other out.

\section{Benchmark on Coloradar Dataset}
We compare \textit{Our} approach to \textit{x-RIO}~\cite{DoerJGN2022} on the Colo-Radar dataset~\cite{kramer2022coloradar}. \textit{x-RIO}, configured with a single radar and without Manhattan assumption, required modifying \si{16} radar prefiltering parameters for compatibility.
Notably, \textit{Ours} only needs the Doppler standard deviation, which we kept unchanged.
Both approaches used different \acp{IMU}, so we adjusted the \ac{IMU} parameters to match the ColoRadar dataset.
Barometer support was omitted as it is not provided in the dataset.
We calibrated the gyroscope turn-on-bias for both approaches using each sequence's first \SI{3}{\second}.

We evaluate \ac{RPE}~\cite{grupp2017evo}, split into translational and rotational \ac{RMSE}, and the final XY-Position and yaw drift normalized by distance. 
\reftab{tab:benchmark} shows similar performance between the approaches. 
\textit{x-RIO} slightly outperforms in rotational \ac{RPE} but exhibits strong terminal yaw drift in most datasets.
We suspect their precise binary outlier filtering leads to better estimates, but any false positive induces inconsistencies, such as a wrong yaw rate.
\textit{Our} system, with smooth outlier rejection and fixed yaw rate bias, performs consistently well.
\begin{table}
    \caption{ColoRadar Dataset Results}
    \centering
    \resizebox{\columnwidth}{!}{%
    \begin{tabular}{cccccc}
    \hline
        \multirow{1}{*}{Dataset} & \multirow{1}{*}{Pipeline}& \multicolumn{2}{c}{RPE (RMSE)} & \multicolumn{2}{c}{Final Drift per Distance} \\ &  & Trans. [\si{\percent}] & Rot. [\si{\degree}] & XY [\si{\percent}] & Yaw[\si{\degree\per\metre}]  \\
        \hline
        hallways\_0  & x-RIO & \textbf{4.41} & \textbf{0.39} & \textbf{2.80} & \textbf{0.031} \\
        ~ & Ours & 4.97 & 0.91 & 4.59 & -0.066 \\
        \arrayrulecolor{lightgray}
        \hline
        arpg\_1 &  x-RIO & 3.36 & \textbf{0.48} & 8.28 & 0.752 \\
        ~ & Ours & \textbf{2.88} & 0.98 & \textbf{2.81} & \textbf{0.096} \\
        \hline
        aspen\_0 & x-RIO & 3.15 & \textbf{0.37} & \textbf{2.47} & 0.288 \\
        ~& Ours & \textbf{3.02} & 1.13 & 3.79 & \textbf{0.038} \\
        \hline
        army\_2 & x-RIO & 3.73 & \textbf{0.41} & 3.97 & \textbf{0.017} \\
        ~ & Ours & \textbf{3.63} & 1.10 & \textbf{0.76} & -0.025 \\
        \hline
        classroom\_0 & x-RIO & 6.92 & \textbf{0.51} & 4.73 & 0.116 \\
        ~ & Ours & \textbf{4.28} & 1.52 & \textbf{1.68} & \textbf{-0.006} \\
        \hline
        outdoors\_0 & x-RIO & 3.19 & \textbf{0.37} & \textbf{1.45} & 0.128 \\
        ~ & Ours & 3.19 & 1.09 & 1.55 & \textbf{0.007} \\
    \arrayrulecolor{black}
    \hline
    \end{tabular}}
    \label{tab:benchmark}
    \vspace{-4mm}
\end{table}

Comparing our dataset and ColoRadar's, our algorithm's final XY-position drift is better in our handheld experiments (\SIrange{0.36}{1.49}{\percent}) than in theirs (\SIrange{0.76}{4.59}{\percent}).
Our hardware likely has a better extrinsic calibration and more suitable, i.e., restrictive, \ac{CFAR} configuration.

\section{Known Limitations}
Despite the robustness and performance demonstrated, the proposed estimator has a few known limitations.
\begin{description}[leftmargin=0pt]
\item[Fixed chirp configuration] limits the radar's range and velocity. To address this, we restricted the quadcopter's maximum input velocity $\gls{control_input}$ to \SI{2}{\metre\per\second} and kept the radar within structure range. Larger velocities or distances would require online chirp adaptation, reducing radial velocity resolution.

\item[Ambient pressure bias changes] are ignored. Abrupt temperature changes, like moving indoors to outdoors, can cause altitude jumps. One solution is resetting the altitude offset or integrating absolute height or temperature measurements.
\item[Extrinsic calibration] is inferred from \acs{CAD}. However, bearing biases still cause noticeable estimation errors. Calibrating additional effects, such as gravity constant, phase biases, or sensor discretization, will enhance future performance.
\item[Yaw rate offset and scale] are hardly observable. Improved yaw observation and vibration management would reduce drift, especially under vibration.
\item[Zero-velocity detection tracking] before take-off and after landing requires at least one repeatable \ac{CFAR} detection. Otherwise, the estimator relies on \ac{IMU} readings and drifts.
\end{description}

\section{Conclusion} 
\label{sec:conclusion}
This work presents an open-source m-estimator for \acl{BRIO}. The estimator is robust to moving objects, ghost targets, and aerodynamic disturbances.
Radar-stabilized quadcopter flights in human-made and natural environments show its general applicability for multirotor navigation.
Accumulated drift as low as \SI{0.5}{\percent} per distance traveled and robustness in visually and geometrically degraded and dynamic environments highlight the potential of radar to replace lidar or vision-based navigation in \acs{GNSS}-denied environments.
A benchmark shows that our m-estimator has fewer tuning variables and performs more consistently than the \textit{x-RIO} Kalman filter, which uses binary outlier rejection.
Future work may investigate \ac{DoA} accuracy, vibration management, and yaw observability.

\bibliographystyle{IEEEtranN}
{\small
\bibliography{references}}

\newpage
{\footnotesize\printunsrtglossary[type=symbols,title={{\normalsize A. List of Symbols}}]}
\section*{\rev{B. Radar configuration}}
\label{sec:appendix_radar_configuration}
\rev{AWR1843AOPEVM radar configuration using the Texas Instruments mmWave SDK~\cite{mmwave2022}.}
\begin{minipage}[h!]{\linewidth}
\begin{lstlisting}[language=Octave,style=mystyle, label={lst:radarcfg}, caption=]
% Created for SDK ver:03.06
% Created using Visualizer ver:3.6.0.0
% Frequency:77
% Platform:xWR18xx_AOP
% Scene Classifier:best_vel_res
% Azimuth Resolution(deg):30 + 38
% Range Resolution(m):0.214
% Maximum unambiguous Range(m):10.95
% Maximum Radial Velocity(m/s):2.56
% Radial velocity resolution(m/s):0.04
% Frame Duration(msec):100
% RF calibration data:None
% Range Detection Threshold (dB):15
% Doppler Detection Threshold (dB):15
% Range Peak Grouping:enabled
% Doppler Peak Grouping:enabled
% Static clutter removal:disabled
% Angle of Arrival FoV: Full FoV
% Range FoV: Full FoV
% Doppler FoV: Full FoV
sensorStop
flushCfg
configDataPort 921600 1
dfeDataOutputMode 1
channelCfg 15 7 0
adcCfg 2 1
adcbufCfg -1 0 1 1 1
profileCfg 0 77 115 7 15 0 0 100 1 64 9142 0 0 30
chirpCfg 0 0 0 0 0 0 0 1
chirpCfg 1 1 0 0 0 0 0 2
chirpCfg 2 2 0 0 0 0 0 4
frameCfg 0 2 128 0 100 1 0
lowPower 0 0
guiMonitor -1 1 1 0 0 0 1
cfarCfg -1 0 2 8 4 3 0 15 1
cfarCfg -1 1 0 8 4 4 1 15 1
multiObjBeamForming -1 1 0.5
clutterRemoval -1 0
calibDcRangeSig -1 0 -5 8 256
extendedMaxVelocity -1 0
lvdsStreamCfg -1 0 0 0
compRangeBiasAndRxChanPhase 0.0 1 0 1 0 1 0 1 0 1 0 1 0 1 0 1 0 1 0 1 0 1 0 1 0
measureRangeBiasAndRxChanPhase 0 1.5 0.2
CQRxSatMonitor 0 3 4 19 0
CQSigImgMonitor 0 31 4
analogMonitor 0 0
aoaFovCfg -1 -90 90 -90 90
cfarFovCfg -1 0 0 10.97
cfarFovCfg -1 1 -2.49 2.49
calibData 0 0 0
sensorStart
\end{lstlisting}
\end{minipage}

\end{document}